\documentclass[runningheads]{llncs}
\usepackage{graphicx}
\usepackage{amsmath,amssymb} 
\usepackage{color}
\usepackage[width=122mm,left=12mm,paperwidth=146mm,height=193mm,top=12mm,paperheight=217mm]{geometry}
\usepackage{bm}
\usepackage{subfig}
\usepackage{multirow}
\usepackage{makecell}
\usepackage{amsfonts}
\begin{document}
\pagestyle{headings}
\mainmatter
\def\ECCV18SubNumber{374}  
\title{Generative Adversarial Forests for Better Conditioned Adversarial Learning} 



\author{Yan Zuo*, Gil Avraham*, Tom Drummond}
\institute{{\small *Authors contributed equally} \\
{\tt\small \{yan.zuo, gil.avraham, tom.drummond\}@monash.edu}, ARC Centre of Excellence for Robotic Vision, Monash University}

\maketitle


\begin{abstract}
In recent times, many of the breakthroughs in various vision-related tasks have revolved around improving learning of deep models; these methods have ranged from network architectural improvements such as Residual Networks, to various forms of regularisation such as Batch Normalisation. In essence, many of these techniques revolve around better conditioning, allowing for deeper and deeper models to be successfully learned. In this paper, we look towards better conditioning Generative Adversarial Networks (GANs) in an unsupervised learning setting. Our method embeds the powerful discriminating capabilities of a decision forest into the discriminator of a GAN. This results in a better conditioned model which learns in an extremely stable way. We demonstrate empirical results which show both clear qualitative and quantitative evidence of the effectiveness of our approach, gaining significant performance improvements over several popular GAN-based approaches on the Oxford Flowers and Aligned Celebrity Faces datasets.
\end{abstract}

\section{Introduction}
Deep neural networks are considered powerful universal approximators~\cite{hornik1989multilayer}; tying them closely to the field of numerical analysis. In numerical analysis, conditioning and stability are often closely associated. From the perspective of deep learning, ill-conditioning is a common reason for slow and inaccurate learning in many backpropagation algorithms~\cite{chang2017reversible}.

The issue of ill-conditioning is especially prevalent within the realm of Generative Adversarial Networks (GANs)~\cite{goodfellow14}, where there often exists a fragility in the training process. GANs perform optimisation on a loss function which tries to find an equilibrium state in an adversarial game between a generator and a discriminator (typically represented by neural networks). The generator and discriminator networks are matched against each other, using a minimax setup to provide each with contrasting training objectives. Typically, network architectures, choice of hyperparameters and weight initialisation need to be carefully considered for some semblance of learning to occur. 

A common issue with GAN-based training is mode collapse~\cite{goodfellow14}, whereby the training equilibrium state can no longer be maintained and either the discriminator or generator network triumphs over its counterpart. Addressing this problem is an active area of GAN-based research - various recent works in the literature have tried to mitigate this by making modifications on the loss~\cite{arjovsky17} or adding regularisation to the model~\cite{radford15,gulrajani17,roth17}. Ultimately, these techniques can all be viewed as an implicit conditioning on the gradients backpropagated which leads to improving the stability of training~\cite{arjovsky17}. Hence, improving the conditioning of a GAN is crucial towards stable learning. In this work, we approach the task of better conditioning a GAN from a different angle, looking towards improving its architecture. More specifically, we identify that the discriminator is the cornerstone that enables the GAN to learn; we look to modify this component such that it is better conditioned and thus improve training stability of the entire framework.

One common element in a traditional discriminator is the fully-connected (FC) linear layer; these are conventionally used as the final layer in a discriminator network where they serve as an interpreter of the features collected by preceding convolution layers. A fundamental weakness of these FC layers is their inability to correctly interpret \textit{non-linear} data; the introduction of the ReLU activation~\cite{krizhevsky12} somewhat alleviated this, but this does not address the inherent flaw of an FC layer's inability to separate highly-complex correlated data. Ideally, we would like an alternative to the FC layer; one which possesses the capability of dealing with non-linear joint distributions. Incidentally, these are properties that are inherent in decision trees; within the vision community, they have built a reputation of having strong discriminating power, allowing them to handle data that is non-linear and high dimensional, thus giving them the capacity to model complex real problems~\cite{caruana06}.

In this paper, we propose a novel method which stabilises training by leveraging the non-linear discriminating power of decision forests~\cite{zuo17} to offer as an alternative discriminator for a GAN. Specifically, we adopt an approach which allows a decision forest to be combined with a DCGAN~\cite{radford15}, unifying the two frameworks and allowing for end-to-end training of the entire model. We call this approach Generative Adversarial Forests (GAF), which offers the following main contributions:
\begin{itemize}
\item We show how poorly conditioned gradients either destabilises learning or stops it altogether. We show that by using decision forests, we can better condition our GAN and subsequently improve training stability (Sections~\ref{ssec:learning_xor} and~\ref{ssec:learning_cifar10}).
\item We propose a novel, end-to-end trainable framework which combines GANs with decision forests (Sections~\ref{ssec:decision_forest_disc},~\ref{ssec:soft_decision_trees} and~\ref{ssec:soft_residual_forest}).
\item We develop a new metric to quantitatively compare the performance of a GAN to other GANs (Section~\ref{ssec:compete-GAN}). We use this metric to demonstrate significant improvements (both qualitative and quantitative) offered by our GAF model over several other popular GAN-based approaches on the Oxford Flowers and Aligned Celebrity Faces datasets (Sections~\ref{ssec:oxford},~\ref{ssec:celeba} and~\ref{ssec:compete-GAN}).
\end{itemize}

\section{Background}
\subsection{Generative Adversarial Networks}
Generative Adversarial Networks (GANs) were first introduced in~\cite{goodfellow14}; they are a member in the taxonomy tree of Generative Models~\cite{goodfellow2016nips} which attempt to derive an explicit estimate of the density distribution~\cite{frey1998graphical,oord2016pixel}. GANs, however, implicitly estimate a data distribution using the following minimax objective loss function:
\begin{equation}
\begin{split}
\min_{G}\max_{D}V(D,G)=E_{x\sim{p_{data}(x)}}[\log{D(x)}] \\ + E_{z\sim{p_{z}(z)}}[\log{(1-D(G(z)))}]
\end{split}
\label{eq:discriminator_loss}
\end{equation}
This loss function narratively describes a game between two opponents, a discriminator and a generator. The discriminator's goal, given an example, is to judge whether that example was drawn from the true distribution $p_{data}(x)$, or from the generated fake distribution $G(z)$. Hence, proper optimisation over the loss function in Eq.~\ref{eq:discriminator_loss} results in an equilibrium state where the discriminator and generator are evenly matched and the discriminator cannot distinguish between real and fake generated samples, assigning a half probability to any sample it receives~\cite{goodfellow14}.

However, issues prevalent with the unstable nature of training these networks due to their minimax loss functions gave rise to several works that attempt to address this issue, either by network architectural solutions or modifications to the loss function. Most notably,~\cite{radford15} proposed an architecture involving deep convolutional networks and regularisation methods such as batch normalisation and empirically showed this to help with stabilising the training of a GAN and delay mode collapse. In~\cite{arjovsky17}, a Wasserstein GAN was proposed which modified the loss function by considering the Earth-mover's distance (EMD), showing how progress of training can be more properly coupled with the evolution of generated samples. Other works offered regularisation techniques which were built around the base architecture of~\cite{radford15} to try and stabilise training by altering its training scheme or adding components around it~\cite{roth17,larsen2015autoencoding,zhao2016energy}.

\subsection{Decision Forests}
A Decision Tree (DT) consists of a set of internal decision nodes and a set of terminating leaf nodes. The internal nodes, $\mathcal{D}=\{d_0, \cdots, d_{N-1}\}$, each hold a decision function $d(\bm{x}; \theta)$, where $\theta$ are the parameters of the decision node. Each decision node performs a hard routing of an input to its corresponding left or right child decision node according to $d(\bm{x}; \theta) : \mathcal{X} \rightarrow [0, 1]$. Collectively, the decision nodes map an input sample, $\bm{x}$, from the root node to one of the terminating leaf nodes: $\ell = \mathcal{D}(\bm{x},\Theta)$, where $\Theta$ are the collected parameters of the decision nodes of the tree. Leaf nodes hold a set of real values, $\bm{q}$, which are formed from the training data:   
\begin{equation}
q(\ell) = \frac{\sum_i \delta(\mathcal{D}(x_i)) v_{i}}{n_\ell}
\label{eq:leaf_route_function}
\end{equation}
where $n_\ell$ is the number of samples routed into leaf node $\ell$, $\delta$ provides the routing function for the sample through the tree to leaf $\ell$ and $v_{i}$ is the observed real value for instance $i$.
A decision forest is an ensemble composed of $\mathcal{T}$ number of DTs which produces an averaged output of its trees: 
\begin{equation}
P(\bm{x},\bm{\Theta},\bm{Q}) = \frac{1}{\mathcal{T}}\sum_{t=1}^{\mathcal{T}}
Q^t(D^t(\bm{x},\Theta^t))
\label{eq:forest_output}
\end{equation}
where $Q^t$, $\mathcal{D}^t$ and $\Theta^t$ are the respective values, decisions and parameters of tree $t$, while $\bm{\Theta}$ and $\bm{Q}$ are the collected parameters of \emph{all} trees' decisions and leaf values.

Decision forests are well known for their strong discriminating power~\cite{quinlan86}, although initially they suffered from variance and stability issues and were prone to overfitting~\cite{ho98}. These issues alleviated via various regularisation methods such as randomisation of the feature subspace and bootstrapping~\cite{breiman96,breiman01,geurts06,ho95}. More recent works have focused on training decision forests such that information is interrelated across trees~\cite{bernard12,schulter13b,schulter13}, or better utilising information from a pre-trained forest~\cite{ren15}. Most modern works now utilise decision forests within a deep learning context, either by using decision tree methods to influence the training approach~\cite{ioannou16,richmond2015}, or explicitly incorporating decision trees as part of the core architecture~\cite{zuo17,kontschieder15,bulo14}.
\section{A Better Conditioned Discriminator}
\label{sec:model_conditioning}
\begin{figure}
\centering
 \subfloat[]{%
  \begin{tabular}{c}
  \includegraphics[height=0.2\textwidth]{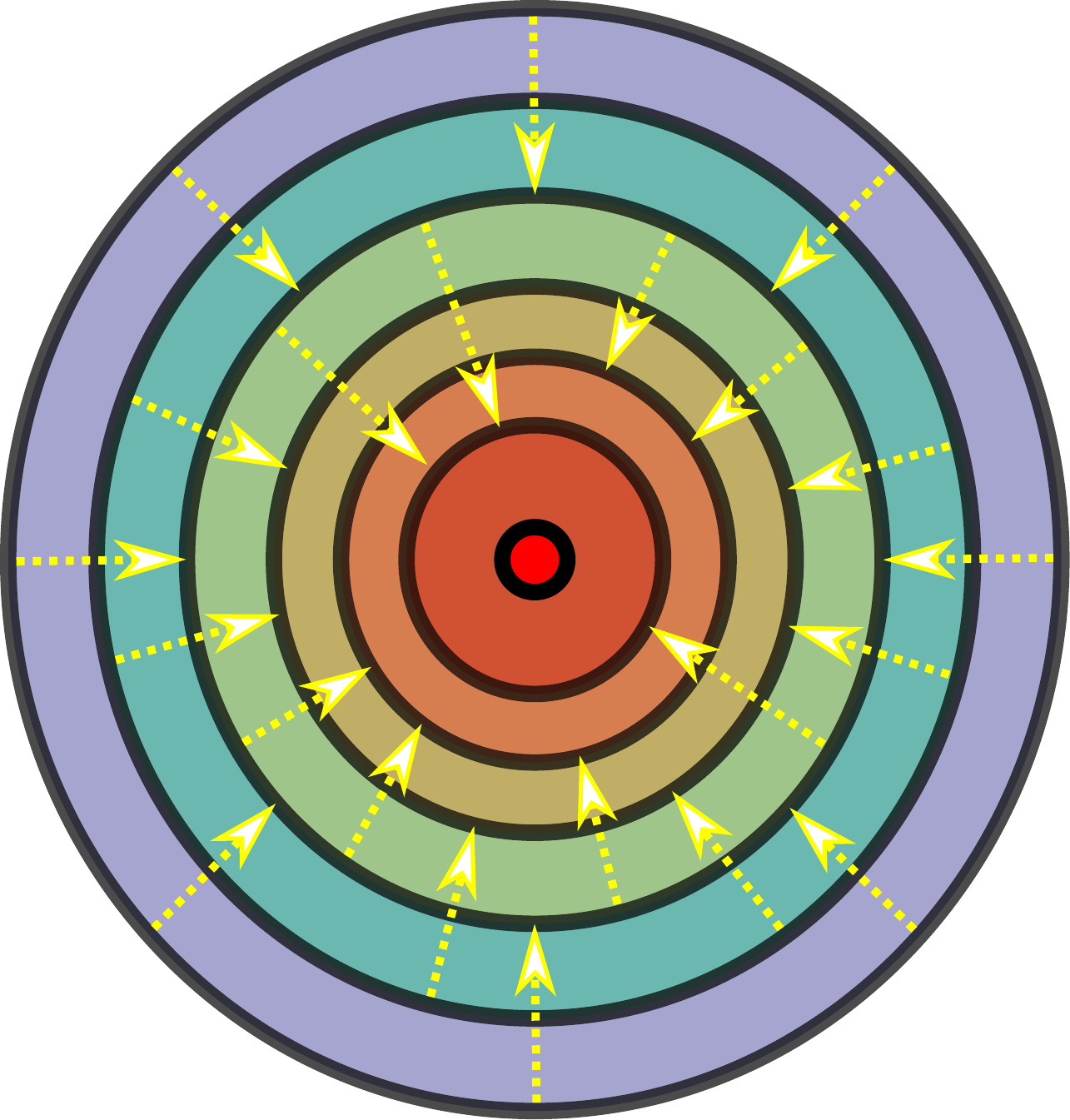}
  \end{tabular}
  \label{fig:good_condition_loss}
 }%
 \hskip 15ex
 \subfloat[]{%
  \begin{tabular}{c}
  \includegraphics[height=0.2\textwidth]{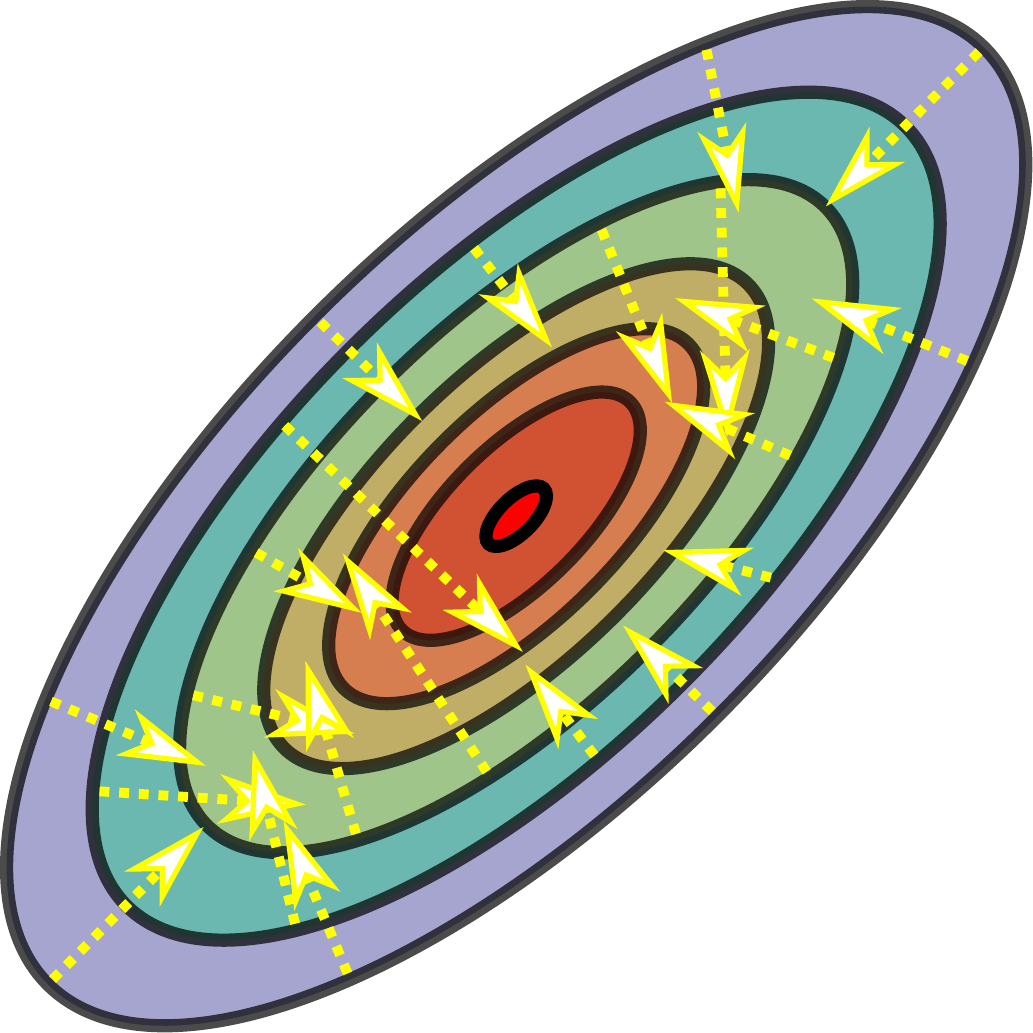}
  \end{tabular}
  \label{fig:bad_condition_loss}
 }%
\vspace{-0.5em}   
\caption{(a): Isotropic loss surfaces of a well-conditioned network. This circular loss surface shows how the minima can be reached from all directions (b): Loss contours of an ill-conditioned model. Due to its non-ideal shape, there are several ways to overshoot and leave the surrounding well, or never enter it.}
\end{figure}

We first take a step back and look at the underlying problem at hand. More importantly, what does it mean for a model to be ill-conditioned? From the perspective of Gradient-Based Optimisation (as is often the case with learning methods), networks which learn in a stable manner should possess an isotropic loss surface which contain local minimas that are surrounded by spherical-shaped wells. These spherical wells allow the minima to be reached from any direction as well as reducing the chance of overshooting the minima. This is illustrated in Fig.\ref{fig:good_condition_loss}. On the other hand, the source of instability in learning can often be traced back to minimas that are surrounded by non-spherical wells~\cite{yuan2008step}; the minimas will be hard to reach and it is easy to overshoot and leave the local area surrounding the minima. Fig.~\ref{fig:bad_condition_loss} illustrates this idea with a ellipsoid-shaped well.

These non-spherical wells around a local minima indicate an ill-conditioning of the model. When a model is ill-conditioned, small perturbations at the input will lead to large changes at the output. This is especially problematic in a GAN framework where a discriminator and generator are playing a minimax game. Any instability in the discriminator updates are subsequently backpropagated to the generator and the amplified result is fed back to the discriminator during the next forward pass.

Turning our attention to the discriminator of a GAN (which is typically a neural network with a FC linear layer), throughout training, the distribution of data that the discriminator (and thus FC layer) sees is complex and joint. This presents an issue: the FC layer's inability to handle non-linear data results in correlated gradients that are backpropagated through the network. Clearly, this is not desirable behaviour as it results in ill-conditioned gradients, leading to unstable learning. With this in mind, replacing the FC layer of the discriminator with an alternative that is capable of separating complex distributions should lead to better conditioned gradients and stabilise training. Here, we look towards decision forests to satisfy this requirement; they possess the non-linearity required to disentangle complex joint distributions in a way FC linear layers cannot. In the following, we will present examples which illustrate this point.
\subsection{Example: XOR}
\label{ssec:learning_xor}
The following example illustrates how a FC linear layer fails to learn an apparently simple 3-dimensional XOR function. This is in contrast to a decision tree which is able to easily learn the function. We construct two models: the first model consists of two FC linear layers with a ReLU non-linearity in between, the second model consists of a FC linear layer connected to a decision tree of depth 2. Across both models, the first FC linear layer consists of 3 hidden nodes which ensures both models have the same modelling capacity. 

\begin{figure}
\centering
 \subfloat[]{%
  \begin{tabular}{c}
  \includegraphics[width=0.32\textwidth]{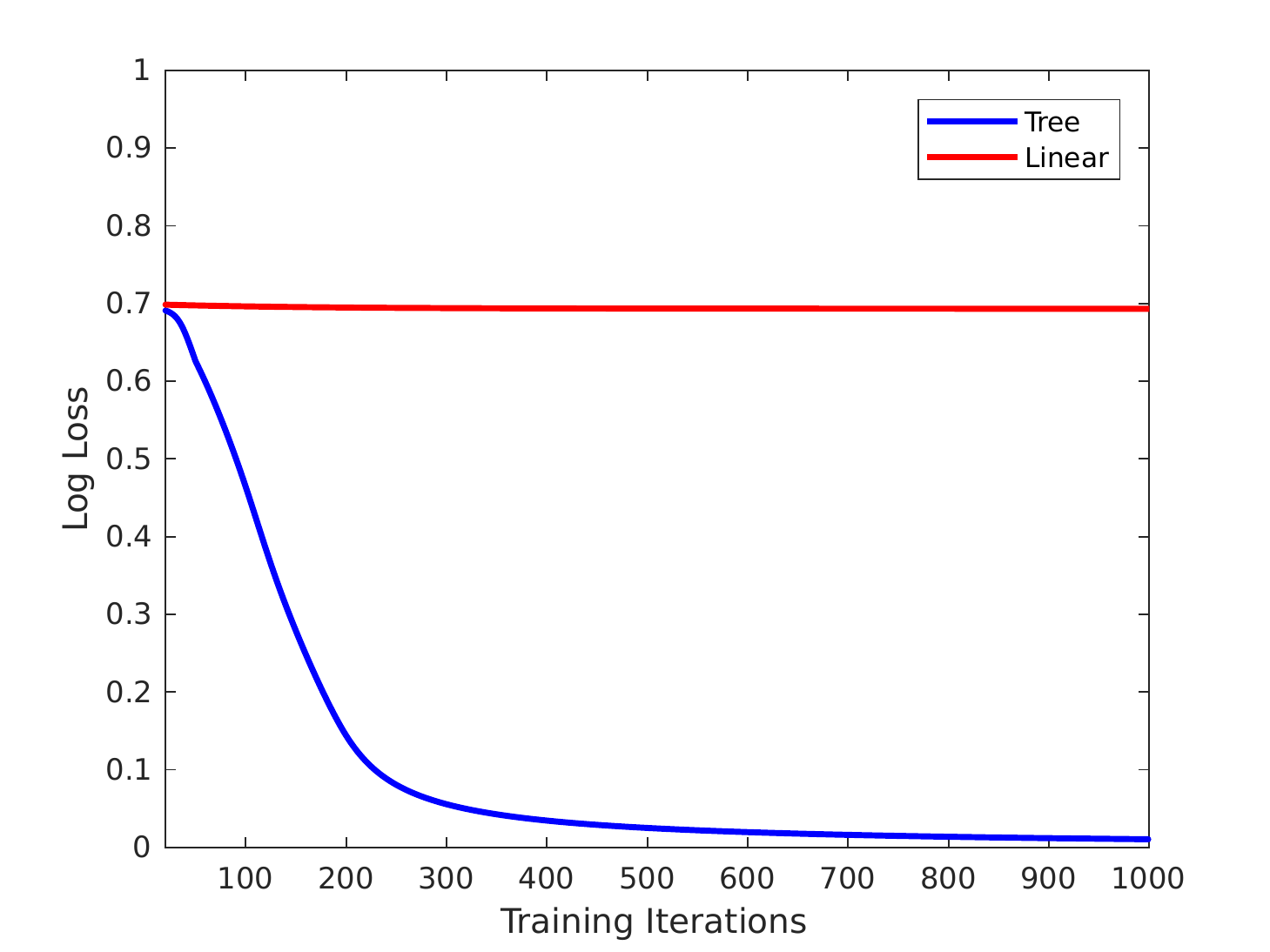}
  \end{tabular}
  \label{fig:xor_loss}
 }%
 \hskip 10ex
 \subfloat[]{%
  \begin{tabular}{c}
  \includegraphics[width=0.32\textwidth]{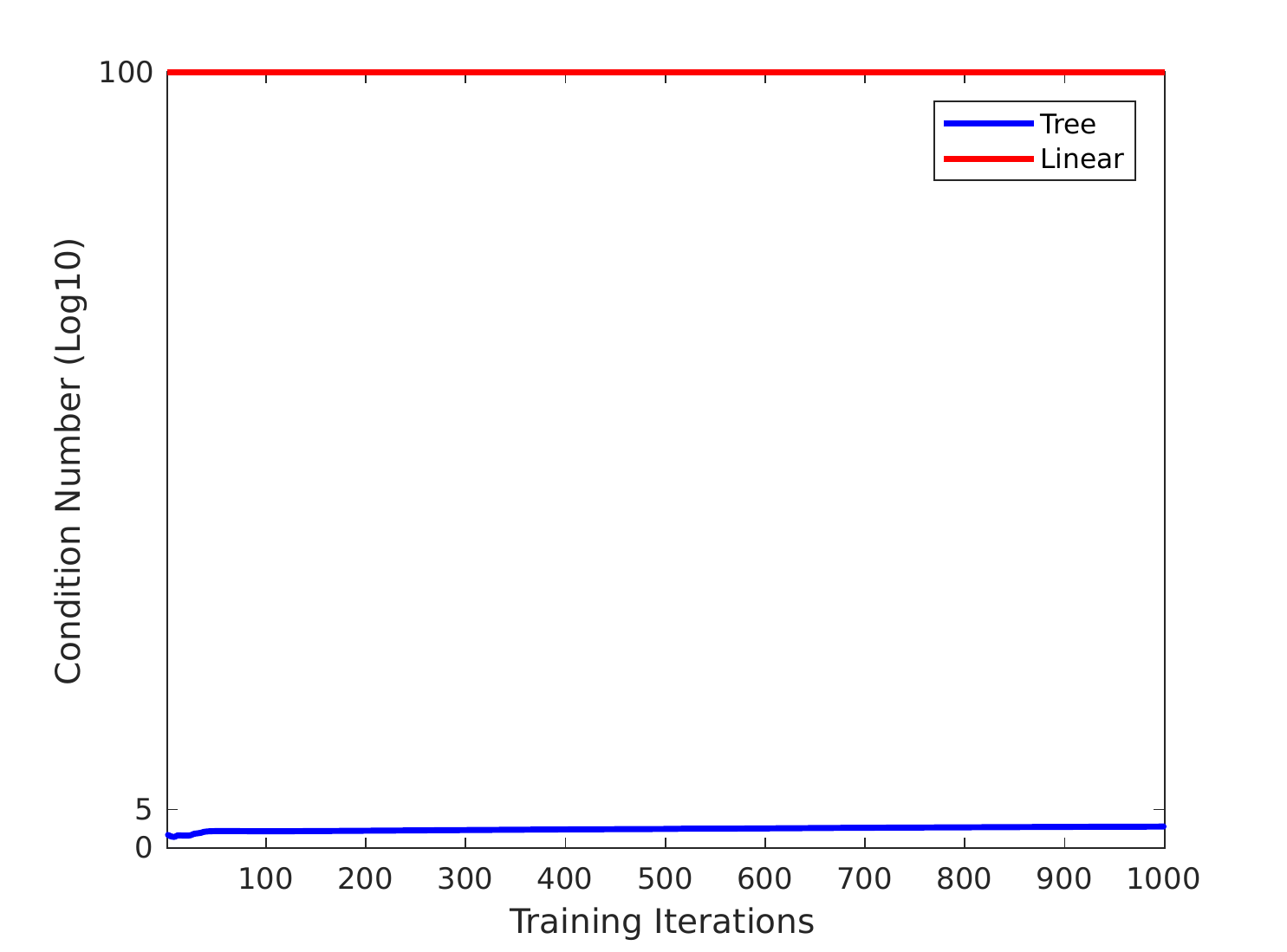}
  \end{tabular}
  \label{fig:xor_condition}
 }%
\vspace{-0.5em}   
\caption{(a): Log loss of 2 FC linear layer model and FC linear + Decision Tree solving a 3-dimensional XOR function. (b): Conditioning of the 2 FC linear layer and FC linear + Decision Tree models, shown on a logarithmic scale.}
\end{figure}

Theoretically, we would expect both models to learn the XOR function. However, in practice we do not observe the model of two FC linear layers with ReLU learning this function; in contrast, the FC linear layer with decision tree is able to learn the function. Fig.~\ref{fig:xor_loss} shows the log loss across 1000 epochs of training for the two models. We can see that the 2 FC linear layer model with ReLU fails to learn the XOR function and resorts to random guessing. The FC linear layer with decision tree successfully learns the XOR function and its log loss quickly converges towards zero.

To explain this behaviour, we look towards the conditioning of the gradients across the two models. We can reformulate the loss as a sum of squares and obtain gradient updates: 
\begin{equation}
\label{eq:least_squares}
\begin{split}
\mathbb{L}(\bm{\Theta}) = &\sum_{i\in\mathcal{B}} \mathbb{L}(i,\bm{\Theta}) = \sum_{i\in\mathcal{B}} \bigg(\sqrt{\mathbb{L}(i,\bm{\Theta})}\bigg)^2 \\
&\implies \Delta\bm{\Theta} = -\bigg[\frac{\partial\sqrt{\mathbb{L}(i,\bm{\Theta})}}{\partial\bm{\Theta}}\bigg]^{\dagger}\sqrt{\mathbb{L}(i,\bm{\Theta})}
\end{split}
\end{equation}
where $\mathbb{L}$ represents the loss, and $i$ are the instances in the batch, $\mathcal{B}$. Eq.~\ref{eq:least_squares} allows us to obtain the condition number of the Jacobian. Fig.~\ref{fig:xor_condition} shows the condition number of the two models. We observe that the FC linear layer model with a decision tree has a much lower condition number compared to its 2 FC linear layer counterpart. This result provides an insight into why the FC linear layer with decision tree is able to learn the XOR function: it is better conditioned and thus provides gradients that reliably decrease the loss function. We observe this result consistently across higher dimensions of XOR.

\subsection{Example: CIFAR10}
\label{ssec:learning_cifar10}
\begin{figure}
\centering
 \subfloat[]{%
  \begin{tabular}{c}
  \includegraphics[width=0.32\textwidth]{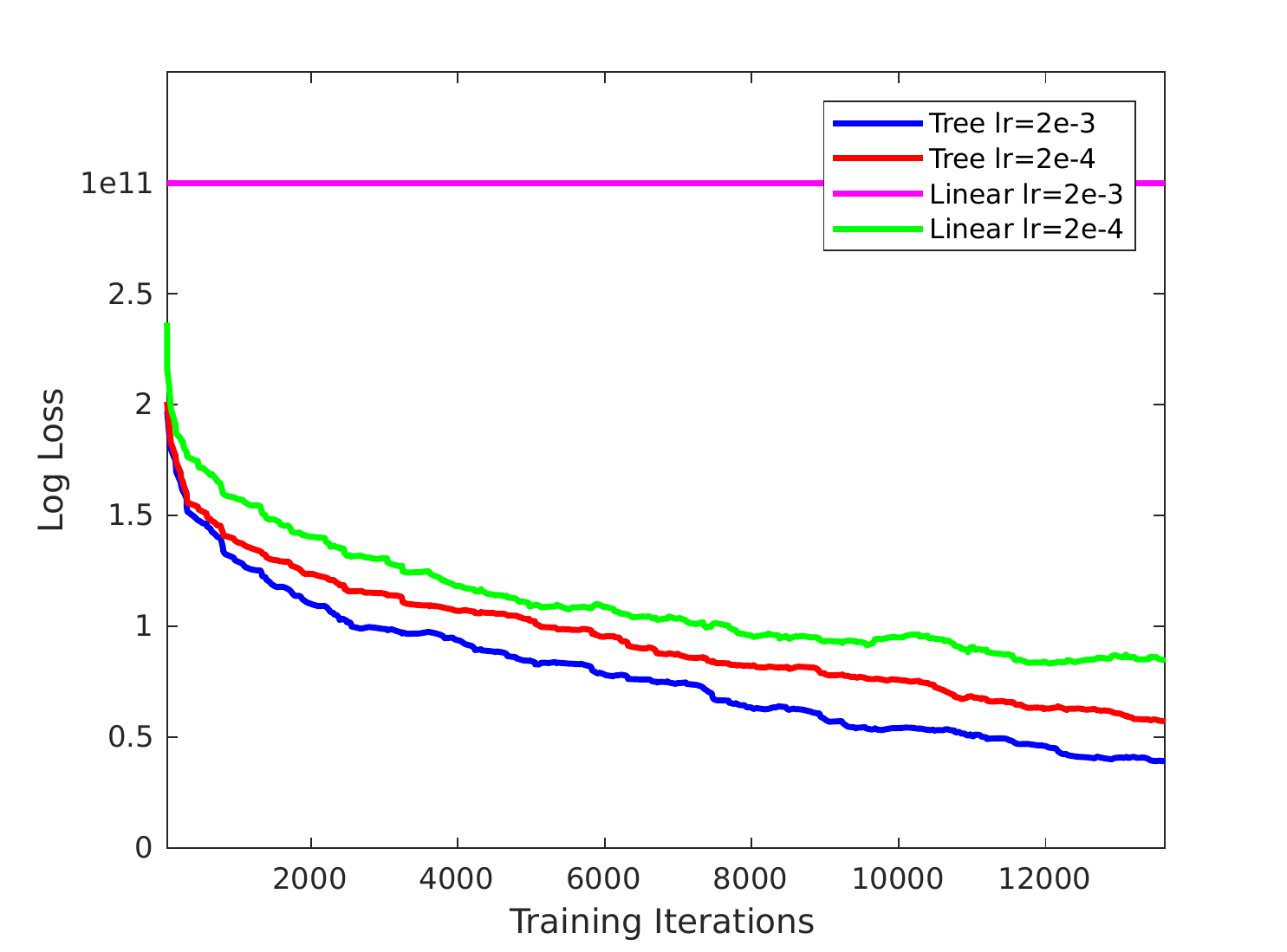}
  \end{tabular}
  \label{fig:cifar_loss}
 }%
 \hskip -1ex
 \subfloat[]{%
  \begin{tabular}{c}
  \includegraphics[width=0.32\textwidth]{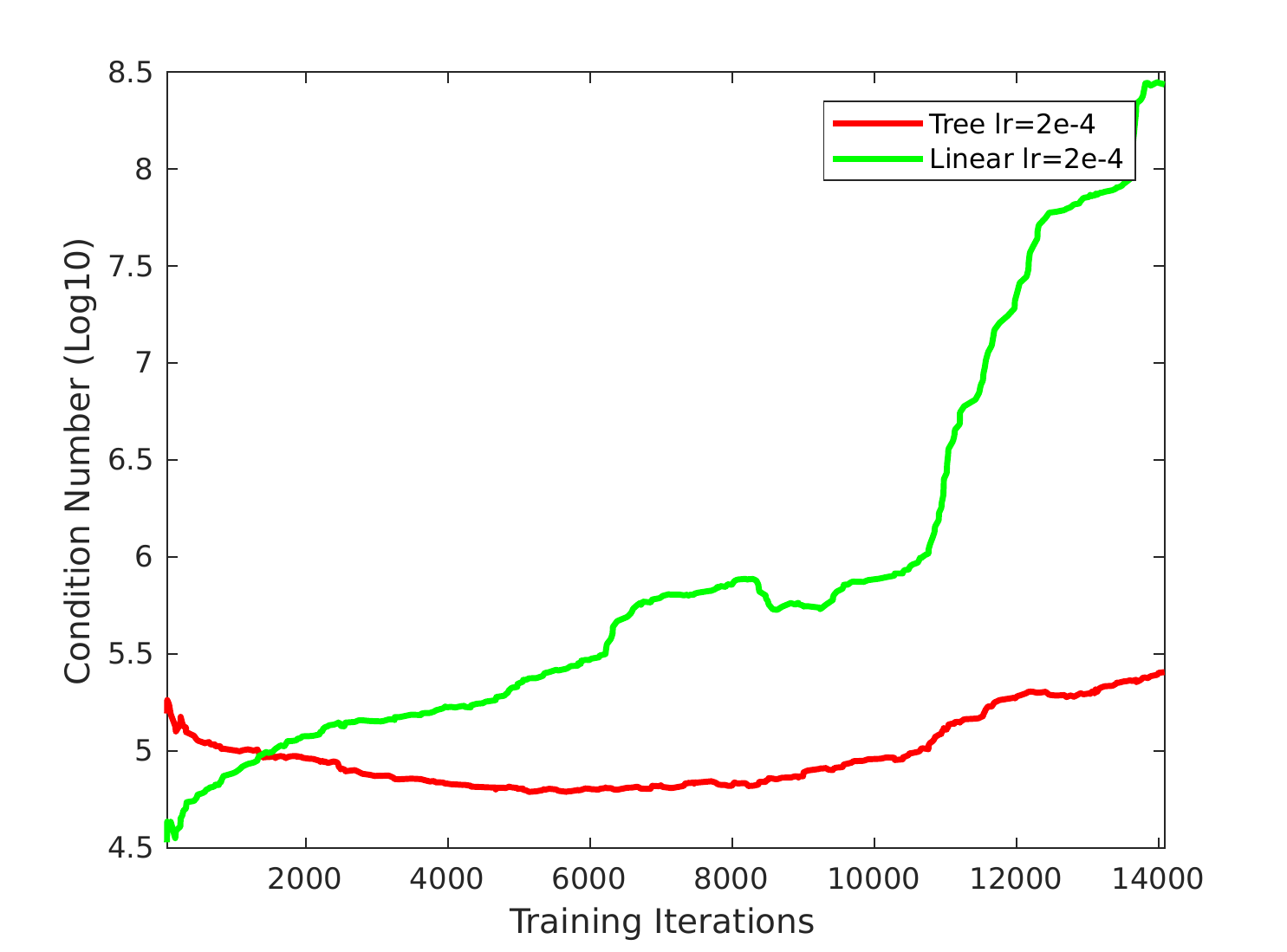}
  \end{tabular}
  \label{fig:cifar_cond_lowlr}
 }%
 \hskip -1ex
 \subfloat[]{%
  \begin{tabular}{c}
  \includegraphics[width=0.32\textwidth]{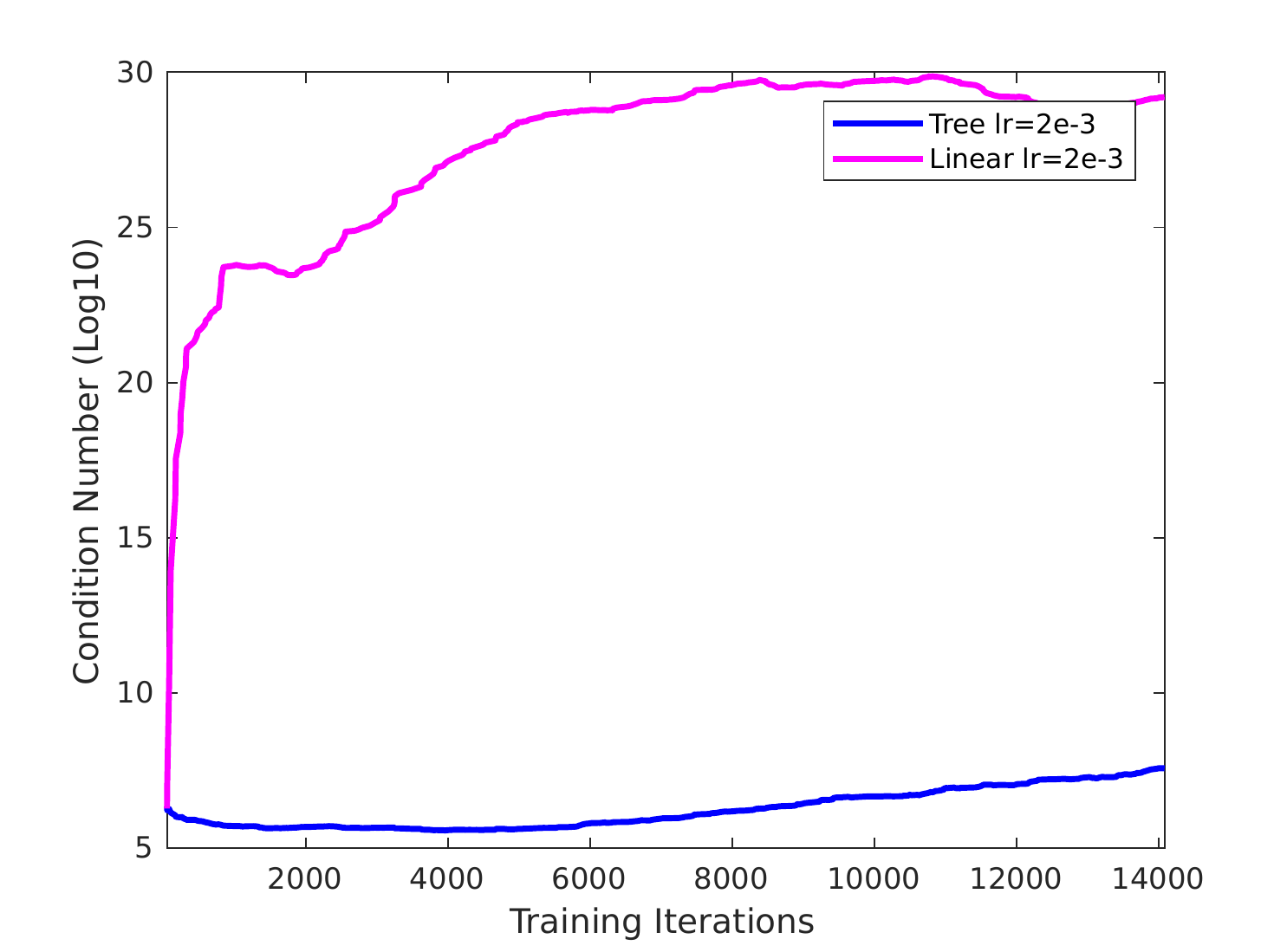}
  \end{tabular}
  \label{fig:cifar_cond_highlr}
 }%
\vspace{-0.5em}
\caption{(a): Log loss of DCGAN discriminator with FC layer and DCGAN discriminator with decision forest for various learning rates. (b),(c): Condition numbers of DCGAN discriminator with FC layer and DCGAN discriminator with decision forest for various learning rates.}
\label{fig:cifar10_overall}
\end{figure}

We propose a second example which demonstrates this property is transferable to real world data. We construct two models for classification on the CIFAR10 dataset~\cite{krizhevsky2009learning}: the first model uses the discriminator in DCGAN~\cite{radford15}; the second model is a modified version of the DCGAN discriminator with the last FC layer replaced with a decision forest (for details on implementation, refer to Section~\ref{sec:generative_adversarial_forests}). We train the DCGAN discriminator for image classification on the CIFAR-10 dataset~\cite{krizhevsky2009learning}, using the settings described in its respective paper~\cite{radford15}. Similarly, we also train the DCGAN discriminator with decision forest on the same task. Fig.~\ref{fig:cifar_loss} shows the log loss of the two models where we can clearly see that the DCGAN discriminator with decision forest converging to a lower loss compared to its counterpart. Similarly, we can use Eq.~\ref{eq:least_squares} to compute the condition number of the gradients backpropagated from the FC layer and decision forest of their respective models. Fig~\ref{fig:cifar_cond_lowlr} shows the conditioning of the two models where we can observe that the conditioning for the discriminator with the decision forest remains relatively well-conditioned compared to the discriminator with the FC layer.

Incidentally, looking at Fig.~\ref{fig:cifar10_overall}, one of the key takeaways from this analysis is that the conditioning of backpropgated gradients allows for points of instability to be identified during the training process. The same cannot be said when merely observing the training log loss. With this insight, we would expect the better conditioned model to learn in a more stable way if we increase the learning rate. Figs.~\ref{fig:cifar_loss}\ and~\ref{fig:cifar_cond_highlr} show the effect of increasing the learning rate by a factor of 10 on both models to their respective log losses and condition numbers. The DCGAN discriminator with decision forest is able to train in a stable manner whereas the vanilla DCGAN discriminator cannot; in this case, both the log loss and condition number reflect this.

We now have the insight that the FC linear layer is not a sensible choice as an interpreter of features for a discriminator. On the other hand, a decision tree provides the necessary non-linearity to properly disentangle correlated features for a discriminator, thus serving as a more sensible choice and better conditioning the network. This leads us to the next section where we introduce our implementation of a discriminator in a GAN framework.

\section{Generative Adversarial Forests}
\label{sec:generative_adversarial_forests}
\subsection{Decision Forest Discriminator}
\label{ssec:decision_forest_disc}
\begin{figure*}
\centering
 \subfloat[]{%
  \includegraphics[height=0.35\textwidth]{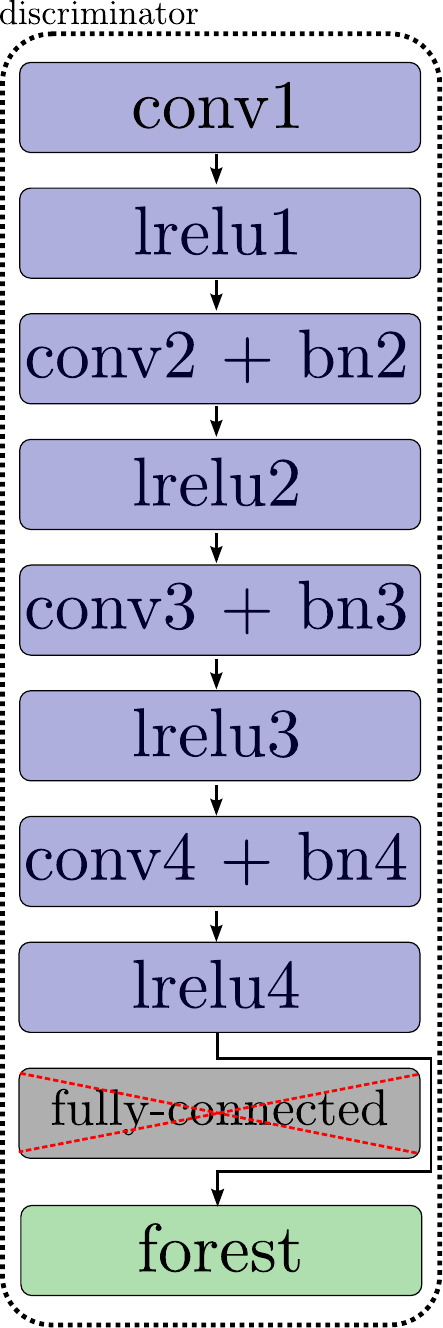}
  \label{fig:discrim_arch}
 }%
 \hspace{2em}
 \subfloat[]{%
  \includegraphics[height=0.35\textwidth]{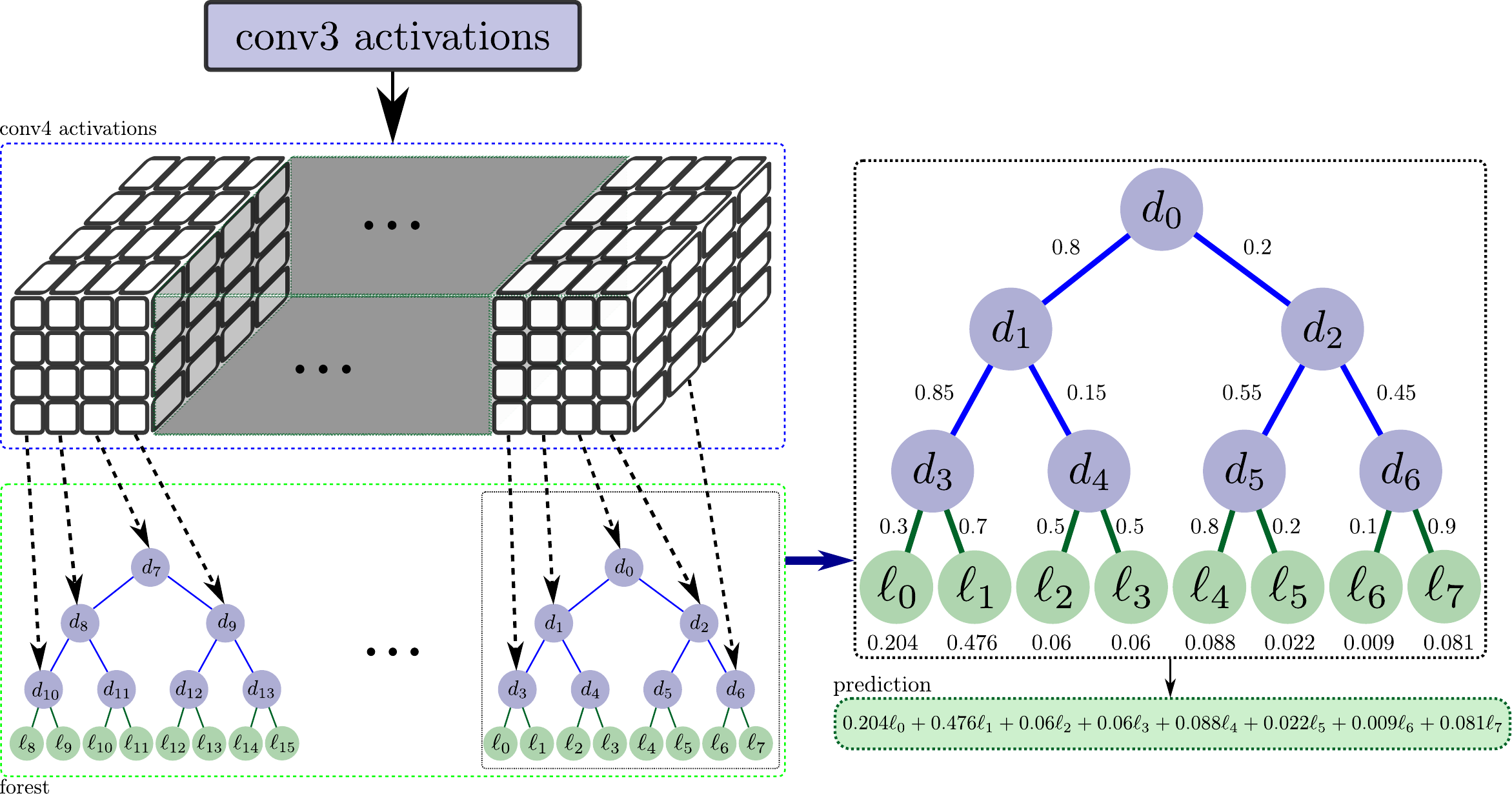}
  \label{fig:conv2trees}
 }%
 \vspace{-0.25em}
\caption{An overview of our proposed changes to DCGAN with a decision forest. (a) shows our architecture which modifies the discriminator network by replacing its fully-connected layer with a decision forest. (b) shows reshaping of \texttt{conv4} activations in DCGAN to form our decision node parameters.}
\label{qual_results}
\end{figure*}

We modify the architecture of the discriminator network in DCGAN~\cite{radford15} by replacing the final fully-connected (FC) layer of the network with a decision forest. This is shown in Fig.~\ref{fig:discrim_arch}. We reformulate the decision nodes in our decision forest such that they are differentiable; hence, the decision forest can be inserted seamlessly into the discriminator network and the whole model can be trained end-to-end. Our method is similar to the approach used in~\cite{kontschieder15} in that we replace the normally hard decision routing function in each decision node with a soft, differentiable sigmoid function. However, we differ in two important aspects: 
\begin{enumerate}
\item We reconstruct the task of learning leaf node values to jointly learn all values in parallel across the ensemble instead of iteratively learning the values.
\item This in turn allows the use of the soft functionality of decision nodes in our ensemble instead of requiring a stochastic hard routing approximation on the forward pass through the trees, as was done in~\cite{kontschieder15}.
\end{enumerate}
In this way, we ensure that the forward pass through our model is consistent with its backward pass and maintain symmetry. Furthermore, our method allows for updating both decision and leaf nodes simultaneously instead of alternating between updates of the two. Our decision forest is used to replace the last FC layer of the discriminator network in a DCGAN~\cite{radford15}. The set of activations output from the last convolution layer (\texttt{conv4}) of the discriminator are reshaped and assigned across the decision nodes in our decision forest (shown in Fig.~\ref{fig:conv2trees}).

\subsection{Soft Decision Trees}
\label{ssec:soft_decision_trees}
Fig.~\ref{fig:conv2trees} shows how each soft decision tree is constructed in our forest. Each tree in the ensemble outputs a single prediction value which is the result of blending the values in all leaves in the tree according to their generated proportion values. 
\paragraph{Soft Decision Functions}
Each decision function in a decision node delivers a value that indicates the proportion of each left and right subtree 
\begin{equation} 
d_n(\bm{x},\bm{\Theta}) = \sigma(\alpha_n (x_n-b_n))
\label{eq:soft_decision_function}
\end{equation}
where $\sigma(x) = (1+e^{-x})^{-1}$ is a sigmoid function with $\alpha_n$ indicating its steepness. $x_n$ and $b_n$ are the respective activation and bias values assigned to decision node $n$. We define $\mu_\ell(\bm{x},\Theta)$ as the blending function which determines the proportion of contribution by leaf $\ell$ towards the tree's final output:
\begin{equation} 
\mu_\ell(\bm{x}|\Theta) = \prod_{n\in\mathcal{N}} d_n(x,\Theta)^{1_{\ell \swarrow n}} \bar{d}_n(\bm{x},\Theta)^{1_{\ell \searrow n}}
\label{eq:blending_function}
\end{equation}
where $\bar{d}_n(\bm{x},\Theta) = 1 - d_n(\bm{x},\Theta)$. $1_{C}$ is an indicator function which equals $1$ when its condition $C$ is met and $0$ otherwise. Hence, the final prediction value generated by a soft decision tree is given by:
\begin{equation} 
Q(\bm{x},\bm{\Theta}) = \sum_\ell \mu_\ell(\bm{x}|\Theta) q_\ell
\label{eq:soft_decision_function}
\end{equation}
Our decision forest changes the role of each output activation of the last convolution layer of the discriminator network; instead of delivering the final prediction, the activation drives the blending proportions output by its assigned decision node. Furthermore, by enforcing our decision trees to make soft decisions which blend leaf values instead of hard routing samples, our model becomes fully differentiable and we are able to easily generate gradients to update our model via backpropagation.
\subsection{Soft Residual Forest}
\label{ssec:soft_residual_forest}
To combine our ensemble of soft decision trees, we adopt the method used in~\cite{zuo17} and create a layer that acts as an ensemble of residual decision trees which are designed to be jointly optimised in parallel to model the underlying input data. Each decision tree in the ensemble contributes a residual value which is combined with all other residual contributions from other trees in the ensemble. This is achieved by multiplicatively combining the predictive contributions from each tree. Hence, for an input sample the forest outputs a final score value, $S$, given by:
\begin{equation} 
S(\bm{x},\bm{\Theta},\bm{Q}) = \prod_{t=1}^{\mathcal{T}}Q^t(D^t(\bm{x},\Theta^t))
\label{eq:multi_forest}
\end{equation}
We modify the residual forest in \cite{zuo17} by adopting a soft approach in both the forward pass and backward passes. Unlike the training schemes in \cite{kontschieder15} and \cite{zuo17}, which relied on alternating between updates of the leaf nodes and split nodes, our modification allows for \textit{true} end-to-end training of the decision forest, where both leaf and split nodes are simultaneously updated.

\section{Experiments}
\label{ssec:Experiments}
For our experiments, we compare our GAF model to three GAN frameworks: DCGAN~\cite{radford15}, ABC-GAN~\cite{susmelj17} and Wasserstein GAN (WGAN)~\cite{arjovsky17}. We use DCGAN~\cite{radford15} as a baseline GAN in which we replace the final fully-connected (FC) layer in its discriminator network with our forest layer. Additionally, we experiment with a shallow and deep variant of our model. The shallow model (GAF-shallow) consists of a forest layer with 8192 trees, each of which is 1 level deep, whilst the deep model consists of a forest composed of 16 trees, each of which is 9 levels deep. 

Specifying forest configurations in this manner allows us to maintain approximately the same number of parameters as the FC layer that is being replaced; the final FC layer in DCGAN consists of $4\times4\times512=8192$ weight values with corresponding $512$ biases, totaling approximately $10,000$ parameters. Our shallow model consists of $8192$ bias values with $8192\times2=16384$ leaf values, totaling approximately $24,000$ parameters. Similarly, our deep model consists of $16\times511=8176$ bias values with $16\times512=8192$ leaf values, totaling approximately $16,000$ parameters. We set $\alpha=1$ and keep all other aspects of the DCGAN network consistent in our model, including the hyperparameters with a batch size of 64. Our weights were normally initialised with zero-mean and standard deviation of 0.02. We used the ADAM optimiser~\cite{kingma2014adam} with a learning rate of 0.0002 and a momentum of 0.9. For settings in our baseline DCGAN, WGAN and ABC-GAN models, we maintain the default settings specified in their respective papers~\cite{arjovsky17,radford15,susmelj17}.
\subsection{Oxford Flowers}
\label{ssec:oxford}
The Oxford Flowers dataset consists of 8,189 images separated into 102 different flower categories~\cite{nilsback06}. We trained our GAF network and DCGAN with parameters described in Section~\ref{ssec:Experiments} for 120 epochs. Additionally, we trained a WGAN following the parameter settings as specified in~\cite{arjovsky17} for 120 epochs. In Fig.~\ref{qual_results}, we show samples of our model's generated images along with the generated images from~\cite{arjovsky17} and~\cite{radford15}. Comparing our set of generated images, we can see our images exhibit no signs of mode collapse, generating a set of diverse flower images. Additionally, the quality of our generated images compared to those of~\cite{arjovsky17} and~\cite{radford15} is considerably higher (note the higher level of detail in our images on the flower petals and cores compared to those generated by~\cite{arjovsky17} and~\cite{radford15}).



\subsection{CelebA Faces}
\label{ssec:celeba}
The Aligned Celebrity Faces (celebA) dataset is a considerably larger dataset consisting of 202,599 images with 10,177 identity categories~\cite{liu15}. We trained our GAF network and DCGAN with parameters described in Section~\ref{ssec:Experiments} for 10 epochs. Additionally, we trained a WGAN following the parameter settings as specified in~\cite{arjovsky17} for 10 epochs. In Fig.~\ref{qual_results}, we show samples of our model's generated images along with the generated images from~\cite{arjovsky17} and~\cite{radford15}. We can again see that our images exhibit no signs of mode collapse with a wide variety of different facial appearances shown. In Fig.~\ref{fig:generator_evo}, we show the evolution of generated faces over time across all models. Similarly, the quality of our generated images compared to those of~\cite{arjovsky17} and~\cite{radford15} is again considerably higher (note the higher level of fine detail such as wrinkles and hair compared to those generated by~\cite{arjovsky17} and~\cite{radford15}).

\begin{figure*}
\centering
 \subfloat[DCGAN~\cite{radford15}]{%
  \begin{tabular}{c}
  \includegraphics[width=0.25\textwidth]{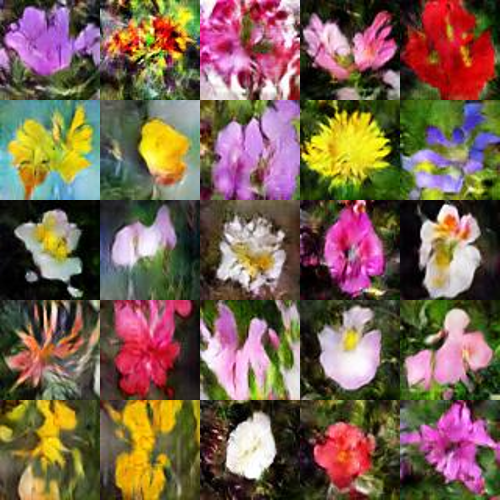} \\
  \includegraphics[width=0.25\textwidth]{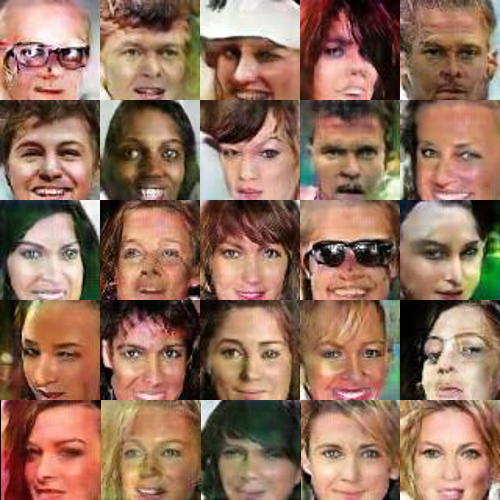} \\
  \end{tabular}
 }%
 \hskip 1ex
 \subfloat[WGAN~\cite{arjovsky17}]{%
  \begin{tabular}{c}
  \includegraphics[width=0.25\textwidth]{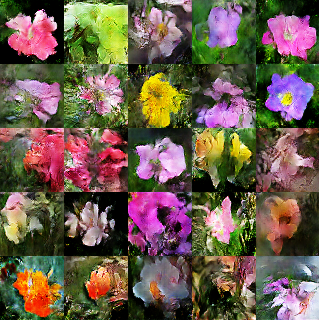} \\
  \includegraphics[width=0.25\textwidth]{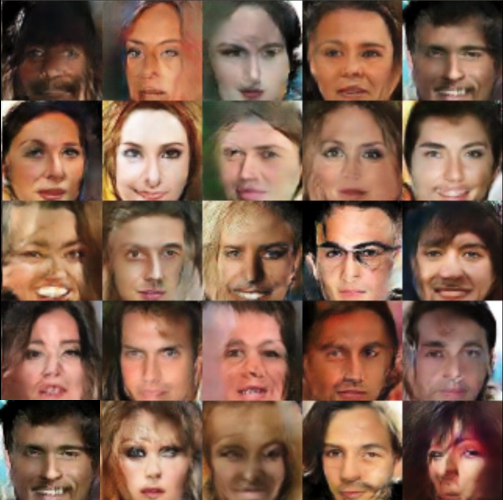}
  \end{tabular}
 }%
 \hskip 1ex
 \subfloat[GAF-deep (Ours)]{%
  \begin{tabular}{c}
  \includegraphics[width=0.25\textwidth]{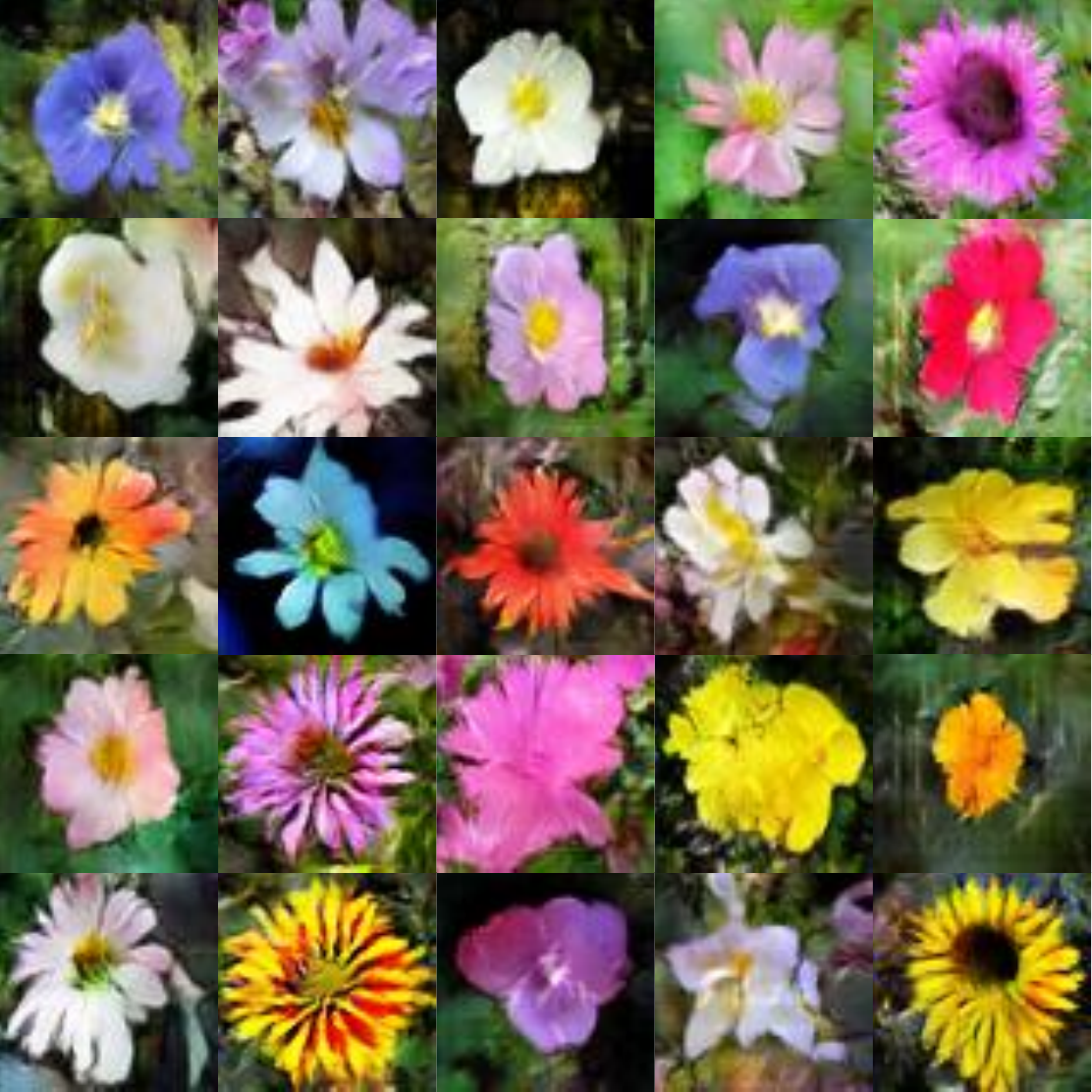} \\
   \includegraphics[width=0.25\textwidth]{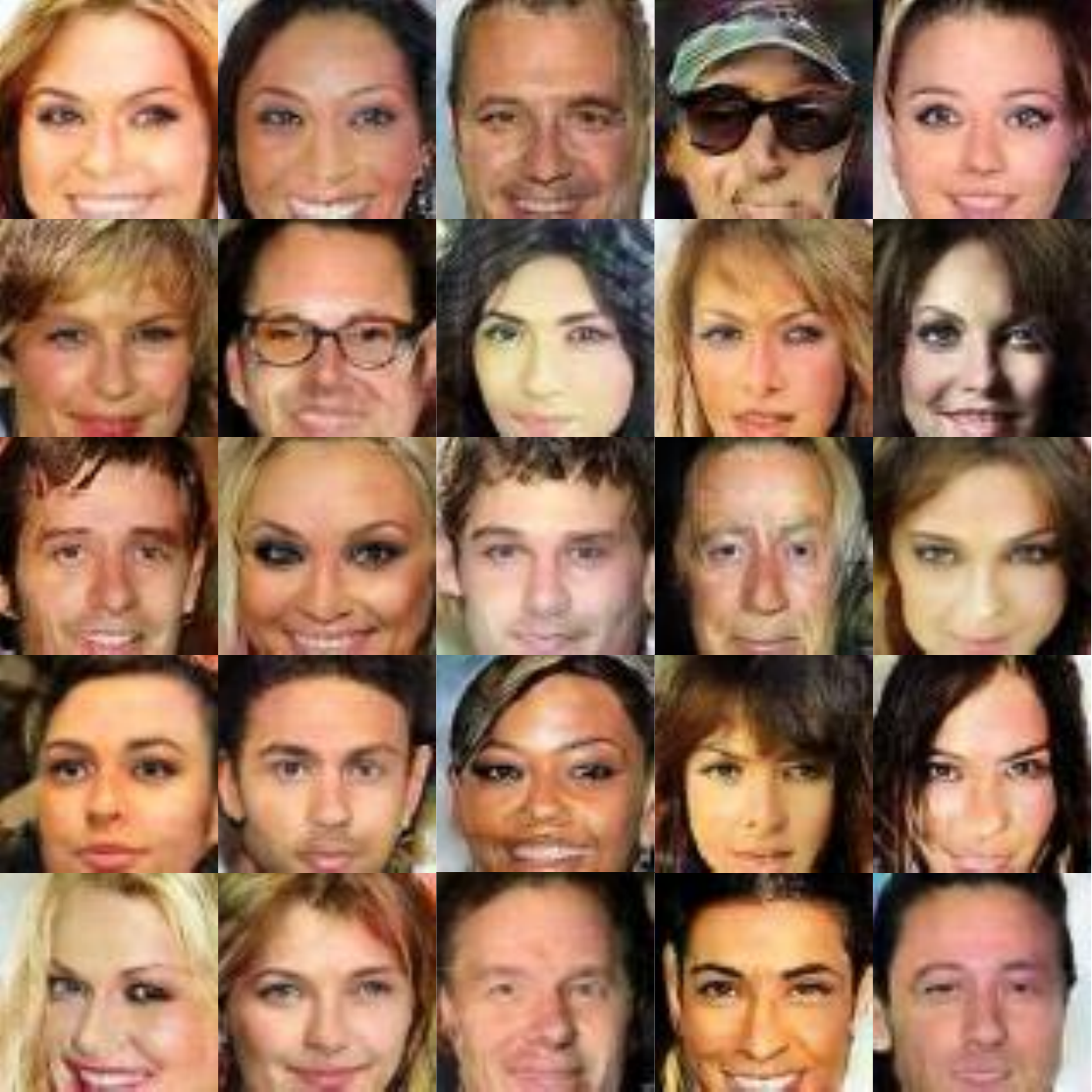}
  \end{tabular}
 }%
 \vspace{-0.25em}
\caption{Qualitative results on Oxford Flowers and celebA datasets. (c) shows our deep GAF model's generator samples. Note the increased level of detail in our generated samples, resulting in sharper looking images.}
\label{qual_results}
\end{figure*}
\subsection{Competitive-GAN Score}
\label{ssec:compete-GAN}
\paragraph{Evaluating GANs}
The most popular method for measuring a GAN's performance quantitatively is Inception Score~\cite{salimans2016improved}, which looks at the $KL$ divergence between a trained Inception Network's learned distribution on some dataset and the distribution of fake generated data from a GAN. However, there is a fundamental flaw in this measure as it does penalise overfitting on the training data~\cite{che2016mode,zhou2017inception}. A GAN which learns to perfectly replicate its training data will score very high on the Inception Score metric. Obviously, this is undesirable behaviour for learning distributions.

Hence, we propose a new method for quantitatively evaluating a GAN called the Competitive-GAN Score. This method addresses the issue of overfitting in GANs and extends upon the methods of~\cite{roth17} and~\cite{im2016generating}. We ensure generalisation in model learning by using a withheld validation set of real images for additional evaluation using a 9:1 split of the entire dataset into respective training and validation sets. Unlike the Generative Adversarial Metric~\cite{im2016generating}, our method allows the withheld validation set to explicitly impact the final score of a GAN. Furthermore, our approach allows us to demonstrate an important transitive property which orders our models from worst to best performance (as we show in Table~\ref{table:transitive_scores})
\begin{figure}
\begin{center}
    \includegraphics[width=0.5\textwidth]{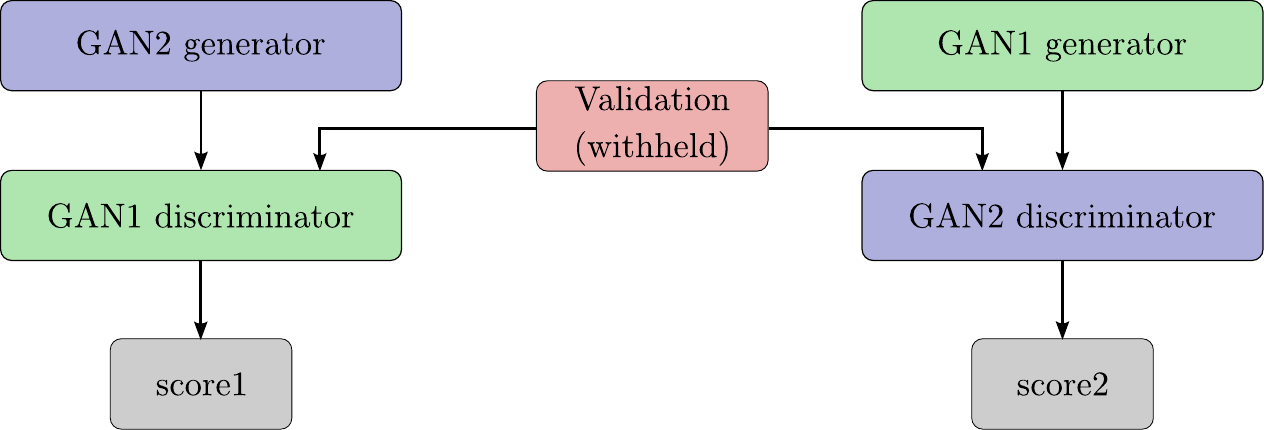}
    \caption{To compare GANs, we pit the generator for each GAN against the discriminator of the other; the additional withheld set of real images punishes GANs that are overfitted}
    \label{fig:gan_wars}
\end{center}
\end{figure}
\paragraph{Generalisation in Learning}
A discriminator which has overfit to its set of real training samples can easily score highly on the metrics proposed in~\cite{roth17} and~\cite{salimans2016improved}; it can spot the other generator's fake samples since they do not belong to the real set of images it has overfit to. Thus, it is able to perfectly classify its own samples as well as generated samples since it has learned a specific set of real images (thereby classifying anything else as fake). Introducing a withheld validation set of real images punishes a discriminator that exhibits this behaviour. A discriminator that is able to perform well on our Competitive-GAN metric indicates a generality in learning - it is able to \textit{generally} distinguish between fake generated samples from real, unseen samples.
\begin{figure*}
\captionsetup[subfigure]{labelformat=empty}
\centering
 \hskip -5ex
 \subfloat[]{%
  \begin{tabular}{c}
  \makebox[20pt]{\raisebox{6pt}{(a)}}
  \includegraphics[width=0.45\textwidth]{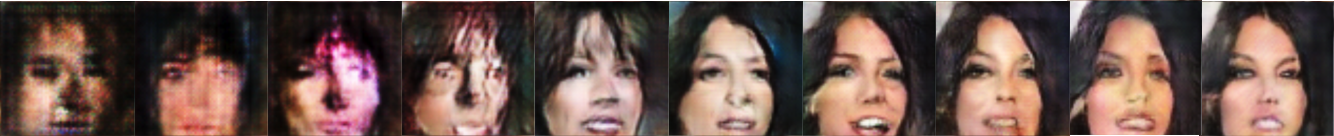} \\
  \makebox[20pt]{\raisebox{6pt}{(b)}}
  \includegraphics[width=0.45\textwidth]{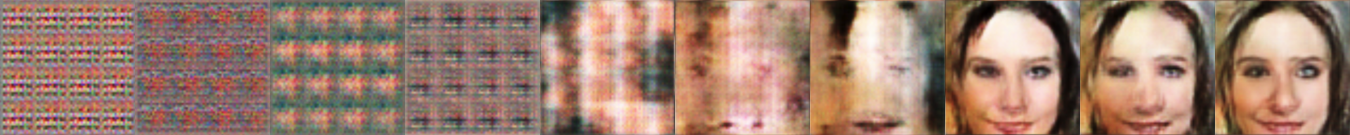} \\
  \makebox[20pt]{\raisebox{6pt}{(c)}}
  \includegraphics[width=0.45\textwidth]{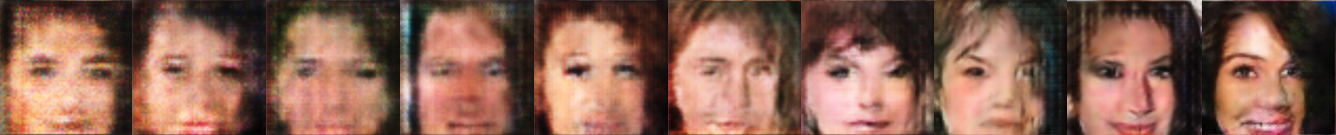}
  \end{tabular}
 }%
 \hskip 1ex
 \subfloat[]{%
  \begin{tabular}{c}
  \includegraphics[width=0.45\textwidth]{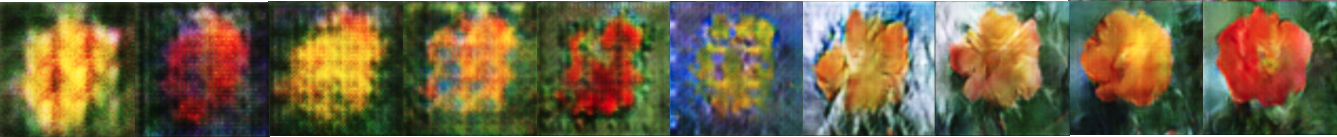} \\
  \includegraphics[width=0.45\textwidth]{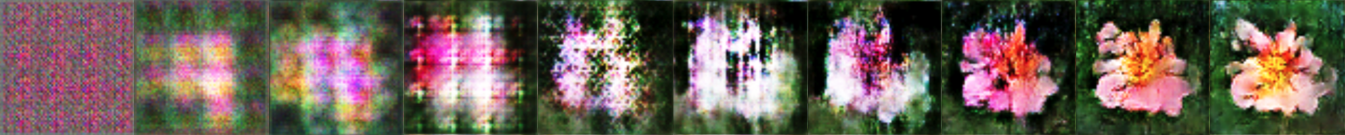} \\
  \includegraphics[width=0.45\textwidth]{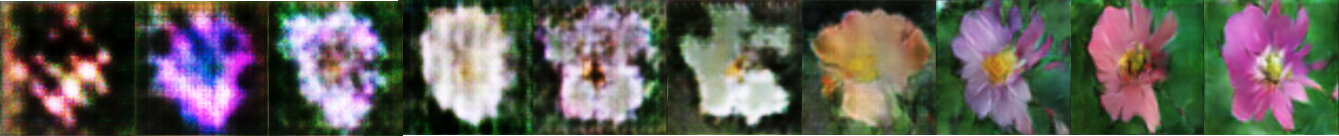}
  \end{tabular}
 }%
 \vspace{-1.25em}
\caption{Evolution of generator output images on the Oxford Flowers and celebA datasets. Row (a) shows the evolution of DCGAN~\cite{radford15}, Row (b) shows the evolution of WGAN~\cite{arjovsky17} and Row (c) shows the evolution of our GAF-deep model. In each case, the evolution over 18k training iterations is shown.}
\label{fig:generator_evo}
\end{figure*}
\paragraph{Adjusted Log Loss}
Using our Competitive-GAN Score metric, each GAN will generate two discriminator log losses; one loss for the indicating performance on unseen generated examples from the other GAN, and one loss indicating performance on the real set of withheld validation images. The validation loss is used to adjust the overall log loss accordingly. Hence, the adjusted log loss, $\mathbb{L}$, of the GAN-A's performance on GAN-B's generated samples, adjusted by validation is given by:
\begin{equation}
\label{eq:adjusted_loss_1}
\begin{split}
\mathbb{L}_{A,B} = &-\frac{1}{N_g}\sum_i \log{(1-D_{A}(G_B(z_i)))} \\
			 &-\frac{1}{N_v}\sum_i \log{D_{A}(x_{v})}
\end{split}
\end{equation}
where $\mathbb{L}_{A,B}$ denotes the discriminator loss for GAN-A cross-evaluated on GAN-B's generated data, adjusted with its log loss on the withheld validation data. $D_A$ is the discriminator of GAN-A, under evaluation, and $G_B$ is the generator of GAN-B, of which samples are being evaluated. $N_g$ is the number of generated samples, $N_v$ is the number of real samples in the withheld validation set. 

In Table~\ref{table:adjusted_losses}, we show the results of evaluating our GAF-shallow and GAF-deep models against DCGAN and ABC-GAN on the Oxford Flowers and celebA datasets (note that the principal diagonal in Table~\ref{table:adjusted_losses} shows the discriminator evaluated against its own generator). We can see that the discriminator of both our shallow and deep GAF models significantly outperform the discriminator of DCGAN and ABC-GAN in adjusted log loss. Due to the different loss function WGAN is trained on compared to the other models, its evaluation was omitted. This comparison would be an unfair one; WGAN's modified loss function would mean its discriminator is specifically trained to work with its own generator; thus it would perform poorly with cross-evaluation of generators from other GANs. However, qualitative comparisons of our models compared to WGAN are shown in Fig.~\ref{qual_results} where GAF clearly offers a significant improvement in quality.

Table~\ref{table:transitive_scores} shows the differences in adjusted log losses between our models. This is done by subtracting our confusion matrix of adjusted losses with the transpose of itself. We can observe a general upwards trend in adjusted loss differences by looking at the upper triangular area of each matrix as we move from DCGAN to our GAF-deep model. This is indicative of a transitive relationship in the ordering of compared GAN models. For the Oxford Flowers dataset, we can see these values indicate that GAF-shallow $>$ GAF-deep $>$ ABC-GAN $>$ DCGAN. Similarly, for the celebA dataset, we observe that GAF-deep $>$ GAF-shallow $>$ ABC-GAN $>$ DCGAN. Note that for the Oxford Flowers dataset, the difference between our GAF-deep and GAF-shallow models is very small (0.03), whilst a significantly larger gap exists between our GAF models compared to ABC-GAN and DCGAN (approximately 1.5).






\begin{table}[h]
\small
\begin{center}
\resizebox{\columnwidth}{!}{
\subfloat[]{\label{table:adjusted_losses}
\begin{tabular}{c|c|c|c|c|c} 
& \multirow{2}{*}{Generator} & \multicolumn{3}{c}{Discriminator} \\ \cline{3-6}
                           & & \shortstack{DCGAN \\~\cite{radford15}} & \shortstack{ABC-GAN \\~\cite{susmelj17}} & \shortstack{GAF \\ (shallow)} & \shortstack{GAF \\ (deep)} \\
\hline
\multirow{3}{*}{\rotatebox[origin=c]{90}{Oxford}} & DCGAN~\cite{radford15} & 3.47 & 2.44 & 1.88 & \textbf{1.78} \\
& ABC-GAN~\cite{susmelj17} & 2.52 & 2.19 & 0.88 & \textbf{0.78} \\
& GAF-shallow (Ours) & 3.26 & 2.62 & 1.63 & \textbf{1.53} \\
& GAF-deep (Ours) & 3.14 & 2.15 & 1.50 & \textbf{1.40} \\
\hline
\multirow{3}{*}{\rotatebox[origin=c]{90}{celebA}} & DCGAN~\cite{radford15} & 3.34 & 2.31 & 2.36 & \textbf{2.03} \\
& ABC-GAN~\cite{susmelj17} & 3.13 & 2.35 & 2.06 & \textbf{1.65} \\
& GAF-shallow (Ours) & 2.86 & 2.34 & 1.65 & \textbf{1.49} \\
& GAF-deep (Ours) & 2.87 & 2.40 & 1.76 & \textbf{1.40} \\
\end{tabular}}
\hspace{2em}
\subfloat[]{\label{table:transitive_scores}
\begin{tabular}{cc|c|c|c|c} 
& & \shortstack{DCGAN \\~\cite{radford15}} & \shortstack{ABC-GAN \\~\cite{susmelj17}} & \shortstack{GAF \\ (shallow)} & \shortstack{GAF \\ (deep)} \\
\hline
\multirow{3}{*}{\rotatebox[origin=c]{90}{Oxford}} 
& \multicolumn{1}{|c|}{DCGAN~\cite{radford15}} & 0 & 0.08 & 1.38 & 1.36 \\
& \multicolumn{1}{|c|}{ABC-GAN~\cite{susmelj17}} & -0.08 & 0 & 1.74 & 1.37 \\
& \multicolumn{1}{|c|}{GAF-shallow (Ours)} & -1.38 & -1.74 & 0 & -0.03 \\
& \multicolumn{1}{|c|}{GAF-deep (Ours)} & -1.36 & -1.37 & 0.03 & 0 \\
\hline
\multirow{3}{*}{\rotatebox[origin=c]{90}{celebA}} 
& \multicolumn{1}{|c|}{DCGAN~\cite{radford15}} & 0 & 0.83 & 0.50 & 0.84 \\
& \multicolumn{1}{|c|}{ABC-GAN~\cite{susmelj17}} & -0.83 & 0 & 0.28 & 0.75 \\
& \multicolumn{1}{|c|}{GAF-shallow (Ours)} & -0.50 & -0.28  & 0 & 0.28 \\
& \multicolumn{1}{|c|}{GAF-deep (Ours)} & -0.84 & -0.75  & -0.28 & 0 \\
\end{tabular}}
}
\vspace{1em}
\caption{(a): Competitive-GAN adjusted losses on Oxford Flowers and celebA datasets: log losses are adjusted to account for performance on an unseen validation set. (b): Difference in Competitive-GAN adjusted losses on Oxford Flowers and celebA datasets. These scores were derived by subtracting Table~\ref{table:adjusted_losses} with its transposed self. Scores are attributed to their corresponding models in the top row (higher is better). A transitive property is observed when inspecting the upper triangular of these tables.}
\label{table:competitive_gan_losses}
\end{center}
\end{table}

\begin{figure}
\begin{center}
 \subfloat[]{%
  \begin{tabular}{c}
  \includegraphics[width=0.32\textwidth]{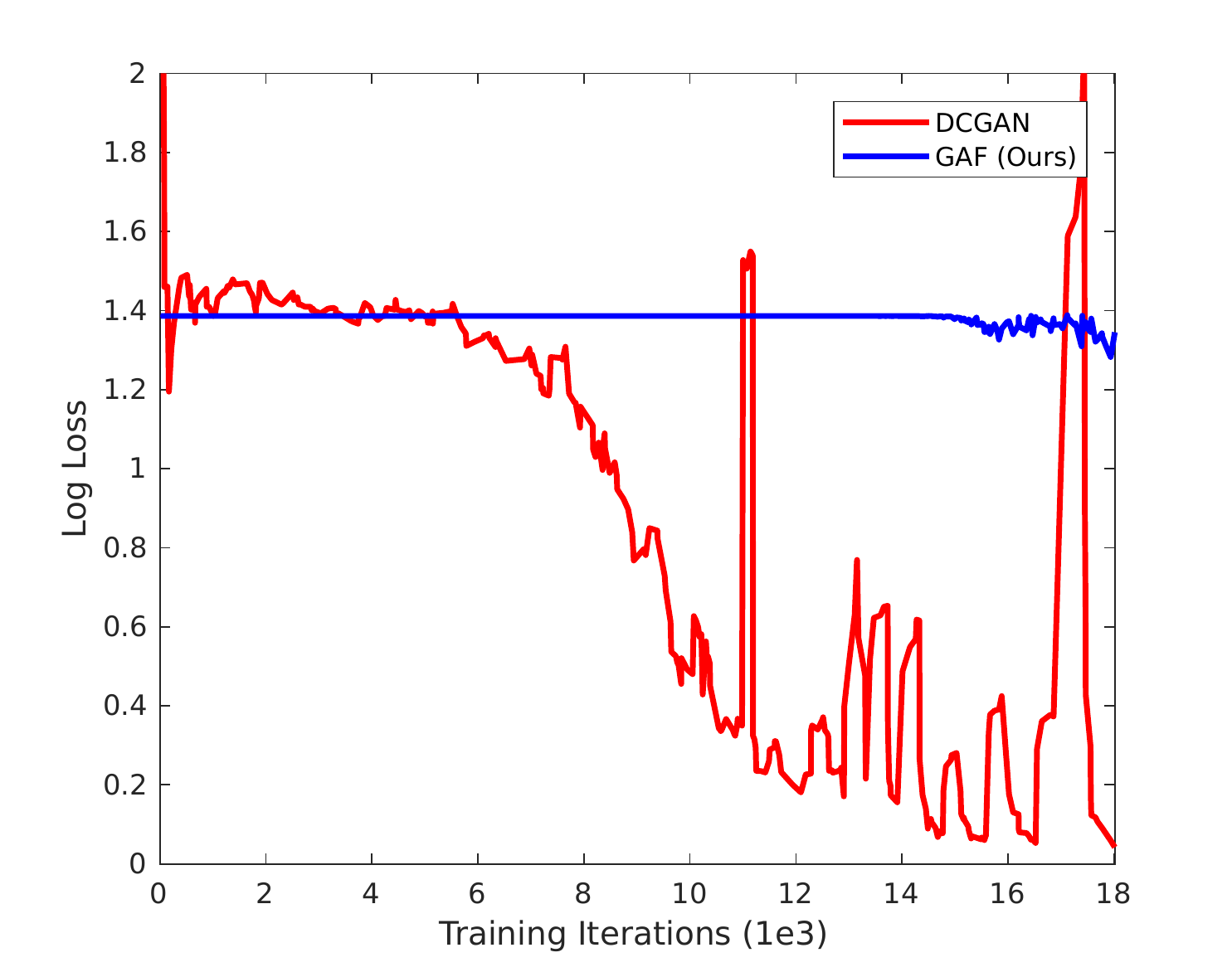}
  \end{tabular}
  \label{fig:train_loss_plot}
 }%
 \hskip -1ex
 \subfloat[]{%
  \begin{tabular}{c}
  \includegraphics[width=0.32\textwidth]{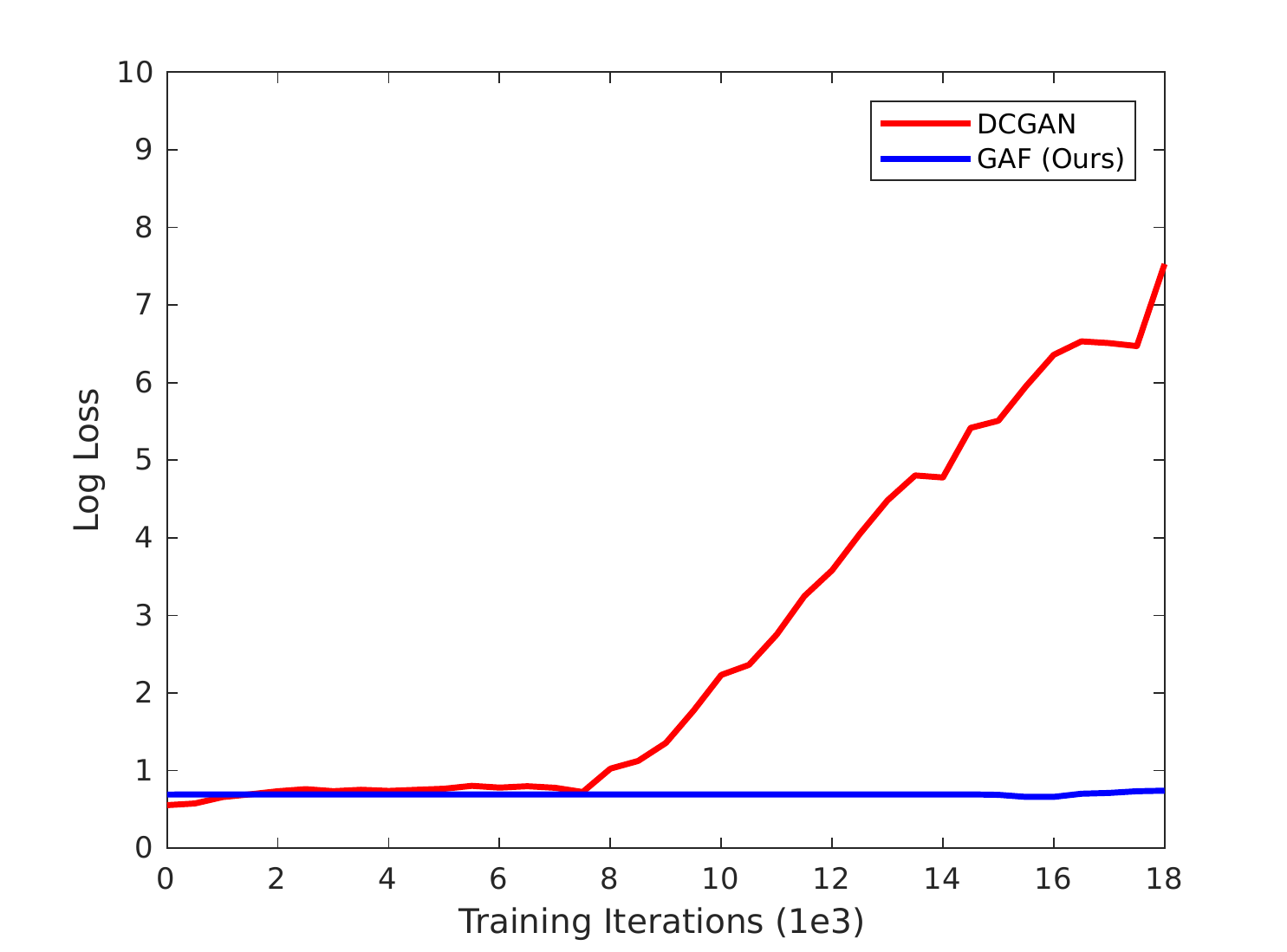}
  \end{tabular}
  \label{fig:val_loss_plot}
 }%
 \hskip -1ex
 \subfloat[]{%
  \begin{tabular}{c}
  \includegraphics[width=0.32\textwidth]{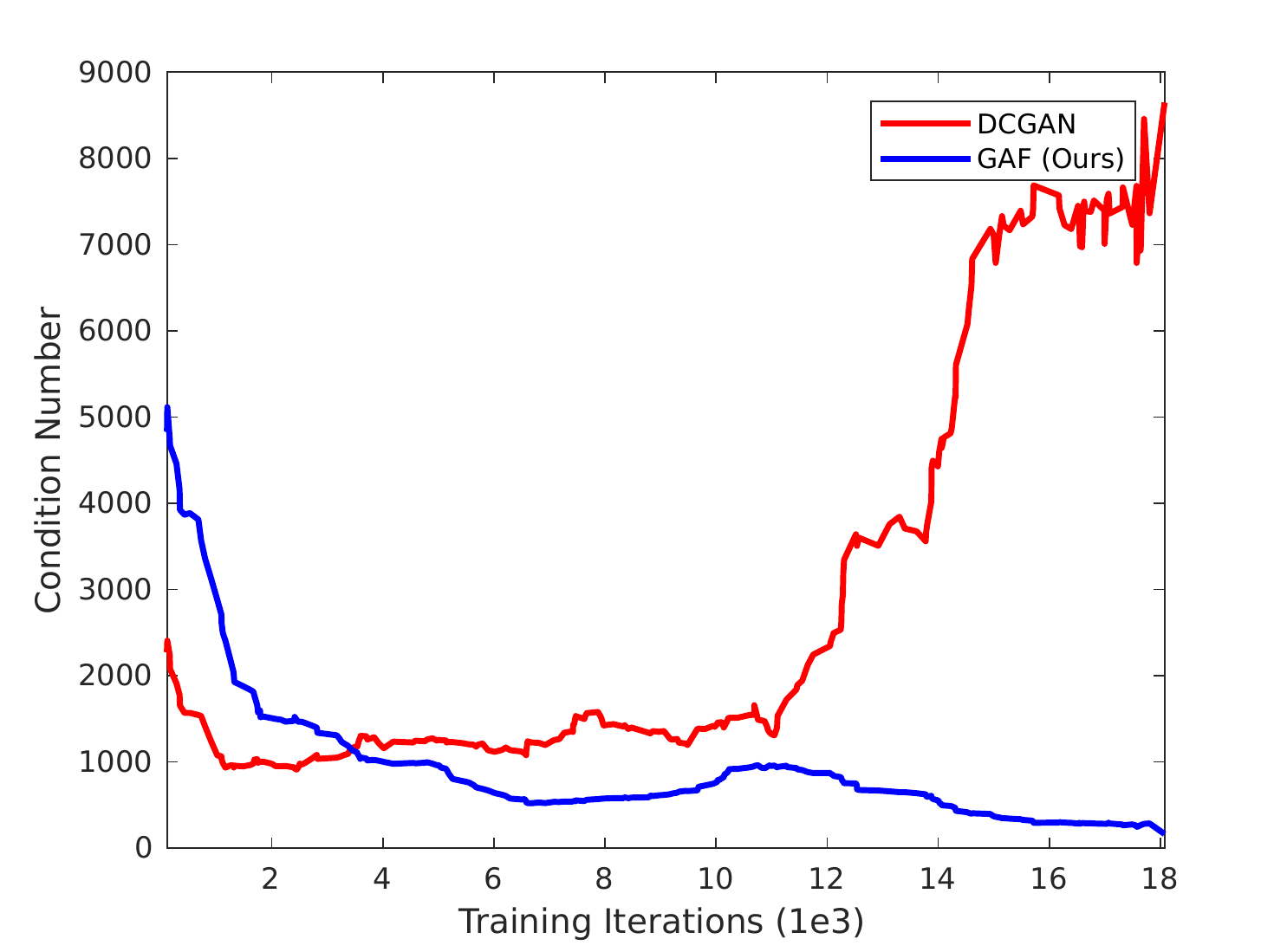}
  \end{tabular}
  \label{fig:conditioning_plot}
 }%
\vspace{-0.5em}    
\caption{(a): Discriminator loss of our GAF model (blue) compared to the DCGAN model in~\cite{radford15} (red). (b): Discriminator loss on the withheld real images for GAF vs. DCGAN comparison. (c) Conditioning of GAF vs. DCGAN.}
\end{center}
\end{figure}

\paragraph{Discriminator Loss Curves} Fig.~\ref{fig:train_loss_plot} shows our GAF model's discriminator log loss compared versus DCGAN's discriminator log loss on the Oxford Flowers dataset. Note that the GAF model's loss signal is significantly more stable than DCGAN's model. Furthermore, we are able to prolong this stability for more than twice the number of training iterations over DCGAN \textit{without} slowing down learning on the generator side (refer to evolution of generated samples in Fig.~\ref{fig:generator_evo}). Looking at the discriminator log loss on the withheld validation set (Fig.~\ref{fig:val_loss_plot}), we can see that our GAF model continues to generalise well past the point where DCGAN begins to overfit. Once again, if we observe the conditioning of GAF compared to DCGAN (Fig.~\ref{fig:conditioning_plot}), we can see that the GAF model remains well-conditioned whilst the conditioning of the DCGAN model deteriorates over time. These results further correlate with the notion that better conditioning of the GAN leads to improved training stability, allowing it to learn in a more general manner.

\section{Conclusion}
This paper presents a new approach for unsupervised training of a GAN, modifying the discriminator with a decision forest. The architecture of our model reflects a key insight that a well-conditioned GAN is crucial to stable learning. We demonstrate that decision forests can provide this necessary conditioning for training stability. Furthermore, we develop a new method for quantitatively measuring the performance of a GAN whilst also placing importance on generalisation in learning. We show why this is important, using our generalised model to show significant improvements both qualitatively and quantitatively over several other GAN-based approaches.

\section*{Acknowledgements}
This research was supported by the Australian Research Council Centre of Excellence for Robotic
Vision (project number CE140100016).

\bibliographystyle{splncs}
\bibliography{egbib}
\end{document}